\documentclass[electronic]{vgtc}             

\graphicspath{{figures/}{pictures/}{images/}{./}} 
\onlineid{1603}
\usepackage{times}                     
\usepackage{mathptmx}                  

\usepackage{amsmath,amssymb}
\usepackage{graphicx}
\usepackage{booktabs}
\usepackage{multirow}
\usepackage{stfloats}                  
\usepackage{placeins}                  

\vgtccategory{Research}

\vgtcinsertpkg

\title{Patient4D: Temporally Consistent Patient Body Mesh Recovery from Monocular Operating Room Video}

\author{Mingxiao Tu\thanks{e-mail:mingxiao.tu@sydney.edu.au} %
\and Hoijoon Jung\thanks{e-mail:hoijoon.jung@sydney.edu.au} %
\and Alireza Moghadam\thanks{e-mail:alireza.moghadam@health.nsw.gov.au} %
\and Andre Kyme\thanks{e-mail:andre.kyme@sydney.edu.au} %
\and Jinman Kim\thanks{e-mail:jinman.kim@sydney.edu.au}} %
\affiliation{\scriptsize University of Sydney}

\teaser{
  \centering
  \includegraphics[width=\linewidth]{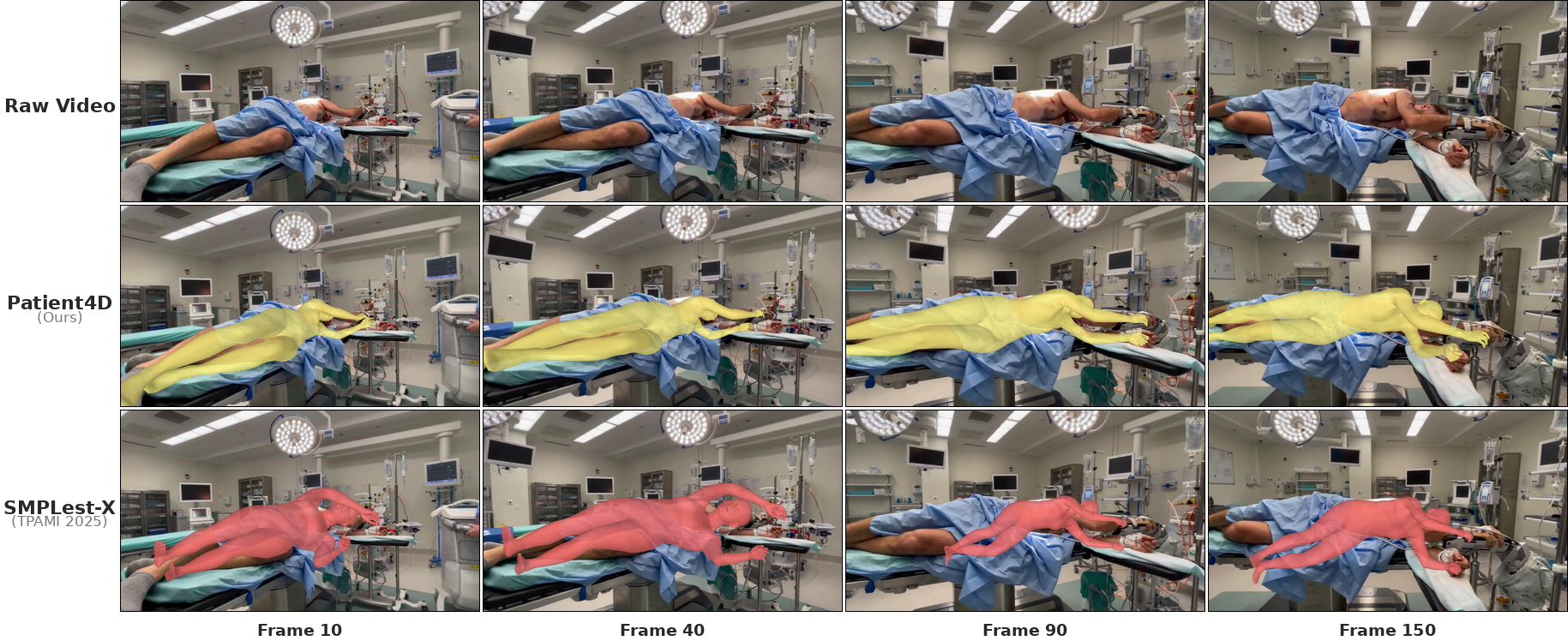}
  \caption{Our proposed Patient4D reconstructs temporally consistent 4D body meshes from monocular surgical video by exploiting the stationarity prior of anesthetized patients. Exemplified on a simulated video of a patient in a lateral position exhibiting high occlusion with limited skin area exposed: as the camera orbits the patient, Patient4D maintains stable mesh alignment across all viewpoints. Top: input raw video. Middle: Patient4D mesh overlays on each frame that correctly estimated the patient body. Bottom: SMPLest-X, a state-of-the-art HMR method, fails to recover the body mesh under complex pose and drape occlusion.}
  \label{fig:teaser}
}

\abstract{
Recovering a dense 3D body mesh from monocular video remains challenging under occlusion from draping and continuously moving camera viewpoints. This configuration arises in surgical augmented reality (AR), where an anesthetized patient lies under surgical draping while a surgeon’s head-mounted camera continuously changes viewpoint. Existing human mesh recovery (HMR) methods are typically trained on upright, moving subjects captured from relatively stable cameras, leading to performance degradation under such conditions. To address this, we present Patient4D, a stationarity-constrained reconstruction pipeline that explicitly exploits the stationarity prior. The pipeline combines image-level foundation models for perception with lightweight geometric mechanisms that enforce temporal consistency across frames. Two key components enable robust reconstruction: Pose Locking, which anchors pose parameters using stable keyframes, and Rigid Fallback, which recovers meshes under severe occlusion through silhouette-guided rigid alignment. Together, these mechanisms stabilize predictions while remaining compatible with off-the-shelf HMR models. We evaluate Patient4D on 4,680 synthetic surgical sequences and three public HMR video benchmarks. Under surgical drape occlusion, Patient4D achieves 0.75 mean IoU under surgical draping, reducing failure frames from 30.5\% to 1.3\% compared to the best baseline. Our findings demonstrate that exploiting stationarity priors can substantially improve monocular reconstruction in clinical AR scenarios.
} 


\begin{document}

\firstsection{Introduction}
\maketitle

\begin{figure*}[!t]
  \centering
  \makebox[\textwidth][c]{\includegraphics[width=1.02\textwidth]{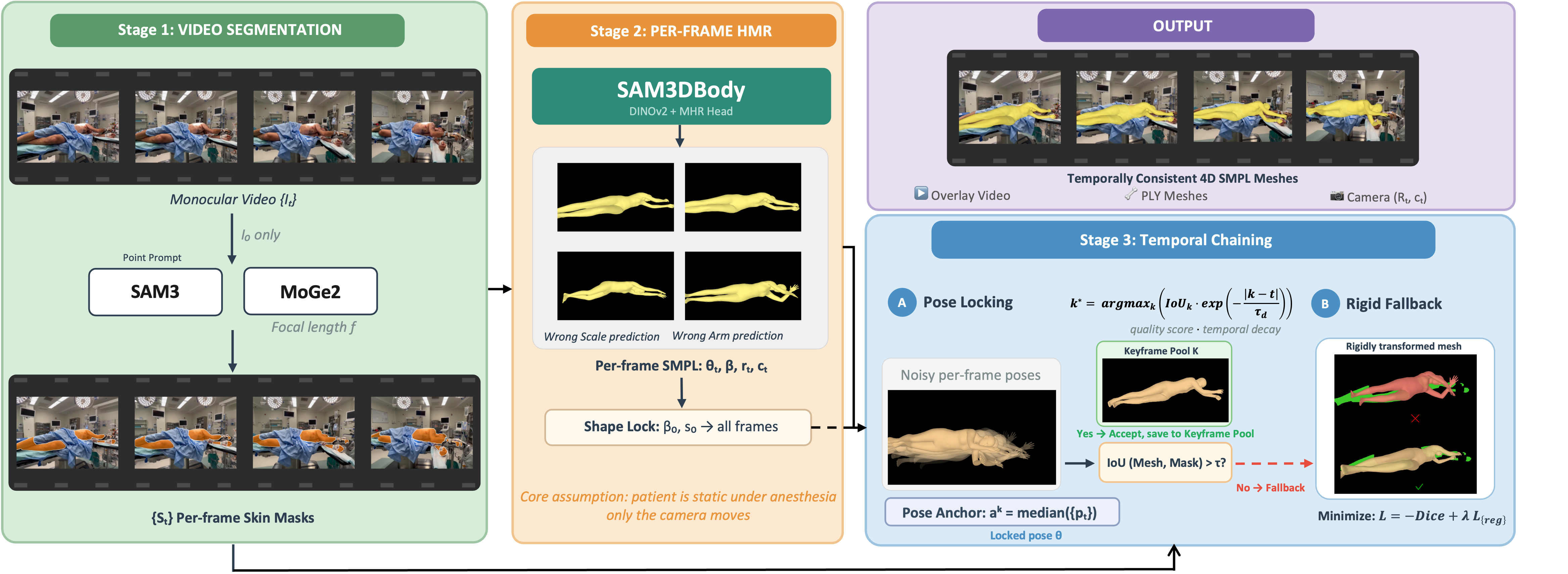}}
  \caption{Overview of the Patient4D pipeline. Given a monocular surgical video and a single point prompt, SAM3 produces per-frame body masks while MoGe estimates camera intrinsics from the first frame. SAM3DBody then recovers 
initial per-frame SMPL meshes from the masked regions; body shape 
$\boldsymbol{\beta}$ and scale are locked to a single keyframe estimate 
and broadcast to all frames. To enforce temporal consistency, Pose Locking 
anchors body pose parameters to a median value calculated across the 
sequence. Frames whose mesh-mask IoU falls below a threshold 
($\tau_\text{IoU}{=}0.6$) are corrected by Rigid Fallback, which selects 
the best reference mesh from a keyframe pool and optimizes a rigid 
transformation to align its projection with the current mask via a Dice 
objective.}
  \label{fig:pipeline}
\end{figure*}

\label{sec:introduction}
Dense 3D human mesh recovery from monocular video has become a fundamental capability for spatial computing and augmented reality (AR) systems. However, most existing methods assume a canonical setting: upright, moving subjects observed from relatively stable viewpoints. In contrast, several critical clinical domains exhibit the exact opposite geometry: in surgical AR, for instance, an anesthetized patient lies motionless beneath heavy surgical drapes while the surgeon's head-mounted camera continuously shifts its viewpoint, yielding only partial views~\cite{liu2024hmr_or, 4dor, labrador}. Despite these challenging visual conditions, accurately recovering the patient's 3D mesh is vital, as it serves as an essential geometric target for registering preoperative CT or MRI~\cite{sttar, neurolens, liebmann2024spine}. Similar static-subject conditions with occlusions appear in sleep monitoring and patient positioning verification~\cite{clever2021bodypressure, pihmr}.

Current systems address this need through dedicated hardware systems. Surgical AR systems commonly use multi-camera setups or optical tracking of rigid fiducial markers attached to the skin~\cite{easyreg, liebmann2019pedicle, vonatzigen2022marker}, while recumbent pose estimation often relies on pressure-sensing mats or thermal arrays. However, these solutions face practical and fundamental limitations. Multi-camera and depth-sensor rigs require fixed mounting and careful calibration that are difficult to maintain in a dynamic OR. Fiducial markers must be affixed before draping, increasing setup time and risking marker falloff~\cite{easyreg}. More fundamentally, line-of-sight sensors are vulnerable to the severe occlusions typical of surgical draping~\cite{Cleveretal2021}. Pressure-sensing mats, while bypassing visual occlusion, yield only sparse skeletal joints rather than the dense meshes required for accurate AR registration. These limitations motivate a monocular RGB approach: the camera already present in the surgeon's head-mounted display requires no additional hardware, calibration, or workflow disruption.

Recent monocular 3D human mesh recovery (HMR) methods have demonstrated high accuracy on moving subjects observed from relatively stable viewpoints~\cite{hmr2, wham, gvhmr, comotion}. Yet applying these methods to the static-subject setting reveals a failure that is not merely a domain gap in pose distribution; it is a structural mismatch in temporal priors. Video-based methods such as GVHMR and WHAM learn motion dynamics from training distributions dominated by upright, and moving subjects~\cite{amass,bedlam}. Their temporal regularizers are designed to smooth plausible human motion trajectories and therefore produce unreliable estimates when the true trajectory is a constant. Per-frame methods avoid this issue but sacrifice temporal consistency, yielding independent predictions that jitter across frames. This motivates a post-hoc mechanism that enforces consistency without modifying the per-frame model itself. The operating room further compounds these tracking challenges: surgical patients lie in complex recumbent poses, including supine, lateral decubitus, and prone, that are largely absent from standard training sets~\cite{disrt_inbed}; surgical drapes create occlusion patterns distinct from those in everyday datasets~\cite{Cleveretal2021}; and the constantly shifting headset camera yields truncated and oblique viewpoints that compound the pose ambiguity. To resolve these issues, the standard approach would be to train or fine-tune a model specifically for the surgical domain. Yet, acquiring clinical 3D ground truth for fine-tuning is difficult due to patient privacy, the intrusiveness of motion-capture rigs during surgery, and the impracticality of instrumenting a sterile environment~\cite{Cleveretal2021}.

To address this, we introduce Patient4D (\cref{fig:teaser}), a stationarity-constrained pipeline that recovers temporally consistent 3D body meshes from monocular surgical video. First, general-purpose vision foundation models segment the exposed body regions that remain visible beneath extensive surgical draping. These 2D masks, combined with an estimated constant camera focal length, are used to generate initial 3D body meshes for each individual frame. While these foundation models yield accurate 3D pose and shape estimates on individual frames, these independent 3D predictions fluctuate erratically when the camera moves. Because the exposed body regions shift constantly with the changing viewpoint, enforcing consistency across the video sequence is essential. To achieve this, we introduce two geometric mechanisms based on the clinical fact that anesthetized patients remain motionless: \textit{Pose Locking}, which anchors the predicted body joint angles to a median value calculated across the sequence to prevent drift, and \textit{Rigid Fallback}, which corrects failed predictions on highly occluded frames by rigidly aligning a previously estimated 3D mesh (from a clearer viewpoint) to the current 2D mask. Together, these mechanisms replace unreliable per-frame predictions and produce temporally consistent 4D reconstructions without requiring surgical 3D training data.

Our primary contributions are summarized as follows:
\begin{enumerate}
    \item We formalize the problem of static-patient human mesh recovery from monocular OR video, where the patient remains static but is occluded with drapes and observed from moving viewpoints with partial views.
    \item We propose Patient4D, a stationarity-constrained pipeline that adapts off-the-shelf, image-level vision foundation models to video sequences, enabling robust 4D human mesh recovery in complex surgical environments without requiring any domain-specific fine-tuning or 3D ground truth.
    \item We introduce \textit{Pose Locking} and \textit{Rigid Fallback} that enforce temporal consistency by exploiting the stationarity prior of anesthetized subjects.
    \item We evaluate Patient4D on 4,680 synthetically rendered surgical sequences designed to reflect real operating room conditions, and three public HMR video benchmarks. Patient4D reduces failure frames under surgical occlusion compared to state-of-the-art baselines, while preserving the accuracy on standard walking subjects. We will publicly release the simulation prompts and generated video sequences to support future research.
\end{enumerate}

\section{Related Work}
\label{sec:related_work}
\subsection{Monocular Human Mesh Recovery}
Monocular 3D HMR has advanced through both image- and video-based approaches. Per-frame methods estimate SMPL~\cite{smpl} parameters from single images, building on advances in transformer architectures~\cite{hmr2}, tokenized pose representations~\cite{tokenhmr}, and mask-conditioned predictions~\cite{prompthmr}. Recently, end-to-end architectures trained on massive, multi-dataset aggregations such as SMPLer-X~\cite{smplerx} and SMPLest-X~\cite{smplesttx}, have set new benchmarks for single-frame accuracy. To ensure temporal consistency, video-based methods incorporate sequential priors: VIBE~\cite{vibe} employs a motion discriminator, WHAM~\cite{wham} and TRAM~\cite{tram} integrate SLAM for world-grounded trajectories, GVHMR~\cite{gvhmr} operates in gravity-aligned coordinates, and CoMotion~\cite{comotion} tracks multiple subjects through occlusion. 

Existing methods are heavily based on training distributions dominated by upright, ambulatory humans, mainly derived from AMASS motion capture and BEDLAM synthetic renderings~\cite{bedlam}. Consequently, the temporal and spatial priors learned from these datasets are mismatched with the non-standard poses typical of surgical patients. For instance, GVHMR's gravity-aligned coordinates assume vertical subjects, and VIBE's discriminator treats supine or lateral configurations as out-of-distribution artifacts. Because deploying motion capture in sterile environments is impractical, clinical 3D ground truth remains unavailable, preventing the resolution of this domain gap through standard fine-tuning.

\subsection{Vision Foundation Models for 3D Perception}
To circumvent the limitations of domain-specific training data, recent work increasingly relies on large-scale vision foundation models that exhibit zero-shot generalization. Models such as the Segment Anything Model (SAM) and its temporal extensions~\cite{sam3} provide robust segmentation under severe occlusion, while monocular geometry estimators like MoGe~\cite{moge} extract reliable depth and camera intrinsics in the wild. Building on these features, recent image-based 3D lifters such as SAM3DBody~\cite{sam3dbody} recover human meshes by lifting 2D segmentation masks and geometric cues into 3D space without requiring dataset-specific pose priors.

However, these foundation models are predominantly designed and optimized for independent, single-frame perception. When applied frame-by-frame to continuous video, minor variations in lighting, occlusion or camera viewpoint can cause severe temporal jitter and topological inconsistencies in the predicted meshes. In the context of surgical AR navigation, where a head-mounted display continuously captures varying viewpoints of a heavily draped patient, these independent per-frame predictions exhibit temporal inconsistency. This instability limits the utility of the resulting meshes for intraoperative registration. Patient4D addresses this limitation by introducing domain-specific temporal mechanisms that enforce structural consistency, effectively adapting these image-level models to provide the stable, continuous 3D patient surfaces required for clinical AR systems.

\subsection{Surgical AR and Patient Registration}
In surgical augmented reality, providing the surgeon with accurate intraoperative navigation requires registering preoperative 3D models (e.g., CT or MRI volumes) to the patient's physical anatomy in the operating room~\cite{sttar, neurolens}. Traditional surgical AR systems achieve this alignment using surface-based registration techniques~\cite{easyreg} or physical fiducial markers. However, these approaches introduce workflow and accuracy limitations in clinical practice. Marker-based registration requires physically attaching fiducials to the patient, which disrupts sterile fields, prolongs preoperative setup, and is vulnerable to line-of-sight occlusions. Furthermore, these physical markers are susceptible to soft-tissue deformation; because they are affixed to the skin, they are highly susceptible to soft-tissue deformation and skin shift between the preoperative scan and the intraoperative posture, leading to degraded registration accuracy. Surface-based registration~\cite{easyreg} matches exposed body regions geometry from a depth sensor to the preoperative model. In the OR, drapes cover most of the body and leave only small skin patches, which makes point-cloud alignment unreliable, because the optimizer can converge to the wrong minimum when the visible surface is geometrically ambiguous.

A separate line of clinical research focuses on estimating patient body pose and shape under occlusions, predominantly for sleep monitoring. However, these approaches typically rely on alternative modalities such as pressure sensing mats~\cite{patientmocap, smpla} or a combination of depth and thermal imaging~\cite{liu2022simultaneously}. These methods need room-specific sensor installations (pressure mats, depth-thermal camera arrays) with their own calibration. Furthermore, purely visual in-bed pose estimators generally predict sparse 3D skeletal joints rather than the dense parametric surfaces required for spatial alignment. Our work bridges these two directions: we reconstruct the patient's continuous 3D SMPL surface directly from the surgeon's monocular RGB camera. By doing so, Patient4D provides a temporally stable 3D surface model that can serve as a direct registration target for preoperative imaging, overcoming the limitations of physical markers without requiring specialized preoperative body scans or multi-modal sensors.
\section{Methodology}
\label{sec:methodology}
Our method is motivated by a key geometric property of operating room capture: the patient remains static while the camera moves around the scene. This observation allows us to treat body pose as a temporally stable variable and enforce global consistency across frames, which forms the basis of the mechanisms described below. Patient4D consists of four stages (\cref{fig:pipeline}). Given a monocular video $\mathcal{V} = \{I_t\}_{t=0}^{T-1}$ of a surgical scene filmed by a moving camera, the pipeline produces a mesh sequence $\{\mathcal{M}_t\}$ parameterized by body pose $\boldsymbol{\theta}$, shape $\boldsymbol{\beta}$, global rotation $\mathbf{r}_t$, and camera translation $\mathbf{c}_t$. Pose and shape are shared across all frames (i.e., $\boldsymbol{\theta}_t = \boldsymbol{\theta}$, $\boldsymbol{\beta}_t = \boldsymbol{\beta}$ for all $t$), reflecting the assumption that the anesthetized patient remained static. The four stages were: (1)~patient segmentation via SAM3~\cite{sam3} (\cref{sec:sam3}), (2)~per-frame mesh recovery via SAM3DBody~\cite{sam3dbody} (\cref{sec:hmr}), (3)~Pose Locking to stabilize body pose across frames (\cref{sec:postprocessing}), and (4)~Rigid Fallback to correct frames where per-frame predictions failed (\cref{sec:rigid_fallback}).

\subsection{Patient Segmentation}
\label{sec:sam3}
SAM3~\cite{sam3} produced per-frame binary masks $\{S_t\}$ of the patient's exposed body regions. A single point prompt on exposed body regions in frame $I_0$ initialized SAM3, which then tracked the segmentation across all subsequent frames via its memory-augmented propagation. In surgical scenes, SAM3 segmented only exposed body regions, excluding drapes and equipment, providing masks that served as the reference for subsequent stages. The pipeline also supported a fully automated mode: when no point prompt was provided, SAM3 segmented all persons in the scene on frame $I_0$, mesh recovery was applied to each, and the patient was identified as the detection whose SMPL spine axis was closest to horizontal, reflecting the lying-down prior. Frames where SAM3 produces an empty mask receive no mesh prediction and are treated as occlusion gaps by the interpolation in the Pose Locking stage (\cref{sec:postprocessing}).

\subsection{Per-Frame Human Mesh Recovery}
\label{sec:hmr}
SAM3DBody~\cite{sam3dbody}, a single-frame HMR model, predicted initial SMPL parameters from each frame $I_t$ conditioned on its corresponding mask $S_t$. Rather than relying on bounding-box crops that include background clutter, SAM3DBody extracts features from the segmented RGB region defined by $S_t$. These mask-conditioned features are then processed by a transformer-based regressor to estimate the 3D body pose $\boldsymbol{\theta} \in \mathbb{R}^{72}$, shape parameters $\boldsymbol{\beta} \in \mathbb{R}^{10}$, and global orientation $\mathbf{r}_t \in SO(3)$. By explicitly guiding the regression with the precise spatial layout of the patient's exposed skin, the backbone mitigates the distraction of complex surgical drapes and instruments. To ensure consistent 3D-to-2D projection across the continuous video sequence, MoGe~\cite{moge} estimated the camera focal length once from the initial frame $I_0$, and this fixed intrinsic value was reused for all subsequent frames.

\begin{figure}[!tb]
  \centering
  \includegraphics[width=\linewidth]{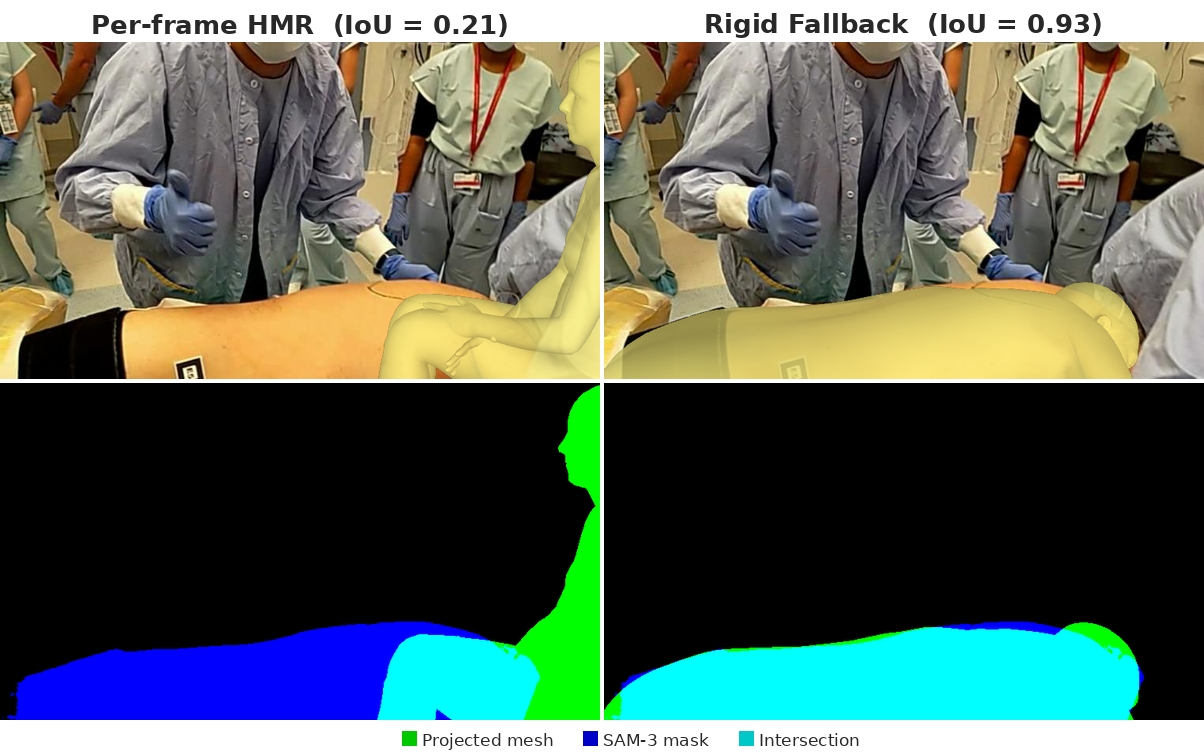}
  \caption{Rigid Fallback on a challenging frame. Left: per-frame HMR produces a misaligned mesh (IoU$=$0.21) when the camera moves to an oblique viewpoint with partial visibility. Right: Rigid Fallback takes the best reference mesh from the keyframe pool and optimizes rotation and camera translation to align its projection with the SAM3 mask via the Dice objective, recovering correct alignment (IoU$=$0.93). Bottom row: mask overlap (green$=$projected mesh, blue$=$SAM3 mask, cyan$=$intersection).}
  \label{fig:rigid_fallback}
\end{figure}

\subsection{Pose Locking}
\label{sec:postprocessing}
Because the patient remains static during surgery, large pose variations predicted across adjacent frames are estimation noise rather than true motion. Pose Locking therefore treats body pose as a globally anchored variable, explicitly constraining frame-wise predictions toward a temporally stable estimate rather than relying on standard temporal smoothing. Before applying Pose Locking, an outlier rejection step filtered erratic predictions, and body shape was fixed to a single keyframe estimate.

\textbf{Outlier rejection.}
For each parameter group $k \in \{\theta, r, \mathbf{c}\}$ (body pose, global rotation, camera translation), frame $t$ was rejected if the L2 distance from the last accepted prediction exceeded a threshold $\tau^k$:
\begin{equation}
d^k_t = \|\mathbf{p}^k_t - \hat{\mathbf{p}}^k_{t-1}\|_2 > \tau^k
\label{eq:outlier}
\end{equation}
Rejected predictions were replaced with the previous accepted value. To prevent indefinite freezing, the pipeline force-accepted after $N_{\max}$ consecutive rejections, blending the current prediction with the last accepted value. This state persisted across processing batches to avoid discontinuities at batch boundaries. Thresholds were set to match the expected inter-frame variation of a static subject at 25~fps; the sensitivity analysis (\cref{tab:sensitivity}) confirms robustness to their exact values.

\textbf{Shape and scale lock.}
Since the patient's body shape did not change throughout the video, the shape vector $\boldsymbol{\beta}$ and body scale $\mathbf{s}$ were fixed to the values from the keyframe pool and broadcast to all frames, preventing per-frame fluctuations in estimated body proportions.

\textbf{Smoothing.}
After outlier rejection, a median filter followed by an adaptive exponential moving average (EMA) was applied. Segments were classified as \textit{static} or \textit{dynamic} based on inter-frame motion magnitude. Static segments received stronger smoothing ($\alpha = 0.01$). Occlusion gaps (frames with empty masks) were filled by linear interpolation from neighboring visible frames. A Kalman filter is the standard alternative, but for a static patient the state transition reduces to the identity. The ablation (\cref{tab:ablation}, row C) confirms that temporal smoothing is counterproductive in this domain; accuracy derives from Pose Locking and Rigid Fallback.

\textbf{Pose Locking.}
This is the key mechanism for temporal consistency. Since the patient does not move, the body pose should converge to a single stable value rather than fluctuating with prediction noise. After a warm-up period of $W$ frames, a pose anchor $\mathbf{a}^k$ was computed as the element-wise median of accumulated predictions:
\begin{equation}
\mathbf{a}^k = \text{median}\big(\{\tilde{\mathbf{p}}^k_t\}_{t=0}^{W-1}\big)
\label{eq:anchor}
\end{equation}
The median was computed element-wise on the axis-angle parameterization of each joint. This does not strictly respect the $SO(3)$ manifold structure. However, after outlier rejection the per-joint variation is small ($<0.3$ rad), where the axis-angle representation is approximately linear and the element-wise median closely approximates the geodesic $L_1$ median on $SO(3)$.

The EMA smoothing target then incorporated a continuous pull toward this anchor:
\begin{equation}
\tilde{\mathbf{p}}_i = \alpha \cdot \mathbf{p}_i + (1{-}\alpha)\big[(1{-}\alpha_a) \cdot \tilde{\mathbf{p}}_{i-1} + \alpha_a \cdot \mathbf{a}^k\big]
\label{eq:pose_lock}
\end{equation}
where $\alpha_a$ is the anchor pull strength. This locked body pose near the true patient pose while allowing global rotation to track the moving camera viewpoint (rotation was smoothed separately in $SO(3)$). The anchor was re-evaluated if segment motion exceeded a reset threshold.

\subsection{Rigid Fallback}
\label{sec:rigid_fallback}
Per-frame HMR predictions remain unreliable when only a small body region is visible, even with temporal smoothing (\cref{tab:ablation}). Since the underlying body configuration does not change significantly, a previously estimated high-quality mesh can serve as a reliable geometric prior. Rigid Fallback exploits this by rigidly transforming a reference mesh to match heavily occluded target frames, solving for the camera viewpoint change (\cref{fig:rigid_fallback}).

\textbf{Formulation.}
Inspired by the memory bank in SAM2~\cite{sam2}, which maintains conditioning frames for robust video segmentation, we adopt an analogous keyframe pool strategy. Rather than fixing a single reference frame, we maintained a \emph{keyframe pool} $\mathcal{K}$ of high-quality predictions, built incrementally during inference. A frame $k$ was added to $\mathcal{K}$ whenever its mesh-mask IoU exceeded a quality threshold $\tau_q$. For each target frame $t$ that required fallback, the best reference was selected from the pool:
\begin{equation}
k^* = \arg\max_{k \in \mathcal{K},\, k < t} \;\text{IoU}_k \cdot \exp\!\big({-|k - t|}/{\tau_d}\big)
\label{eq:keyframe_select}
\end{equation}
where $\tau_d$ controls the temporal decay (in frames), balancing reference quality against proximity. This ensured that the reference mesh $\mathbf{V}_{k^*} \in \mathbb{R}^{N_v \times 3}$ ($N_v$ = number of mesh vertices) was both high-quality and temporally proximate to the target, reducing the magnitude of the required rigid transformation. The first valid frame initialized $\mathcal{K}$.

For each target frame, rotation $\mathbf{R}_t \in SO(3)$ and camera translation $\mathbf{c}_t \in \mathbb{R}^3$ were optimized to align the projected reference mesh with the SAM3 mask $S_t$, minimizing:
\begin{equation}
\mathcal{L} = -\underbrace{\frac{2\,|M_t \cap S_t|}{|M_t| + |S_t|}}_{\text{Dice}} + \lambda_{\text{temp}} \underbrace{\|\boldsymbol{\omega}_t {-} \boldsymbol{\omega}_{t'}\| {+} \|\mathbf{c}_t {-} \mathbf{c}_{t'}\|}_{\text{temporal reg.}} + \lambda_z \underbrace{(c_z^t - c_{z,k^*})^2}_{\text{depth reg.}}
\label{eq:rigid_objective}
\end{equation}
where $M_t$ is the rasterized mesh projection, $\boldsymbol{\omega}_t$ is the axis-angle parameterization of $\mathbf{R}_t$, and $t'$ is the nearest already-fitted frame. We use the Dice coefficient rather than a coverage metric ($|M_t \cap S_t| / |S_t|$) because coverage admits a trivial failure mode: the optimizer can move the mesh closer to the camera, inflating its projection to cover the entire mask without accurate spatial alignment. Dice penalizes the extra pixels that fall outside the mask, requiring spatially accurate alignment. The depth regularization anchored $c_z^t$ near the selected keyframe's depth, preventing a degenerate solution in which the mesh is moved closer to inflate its projection.

\begin{figure}[!tb]
  \centering
  \includegraphics[width=\linewidth]{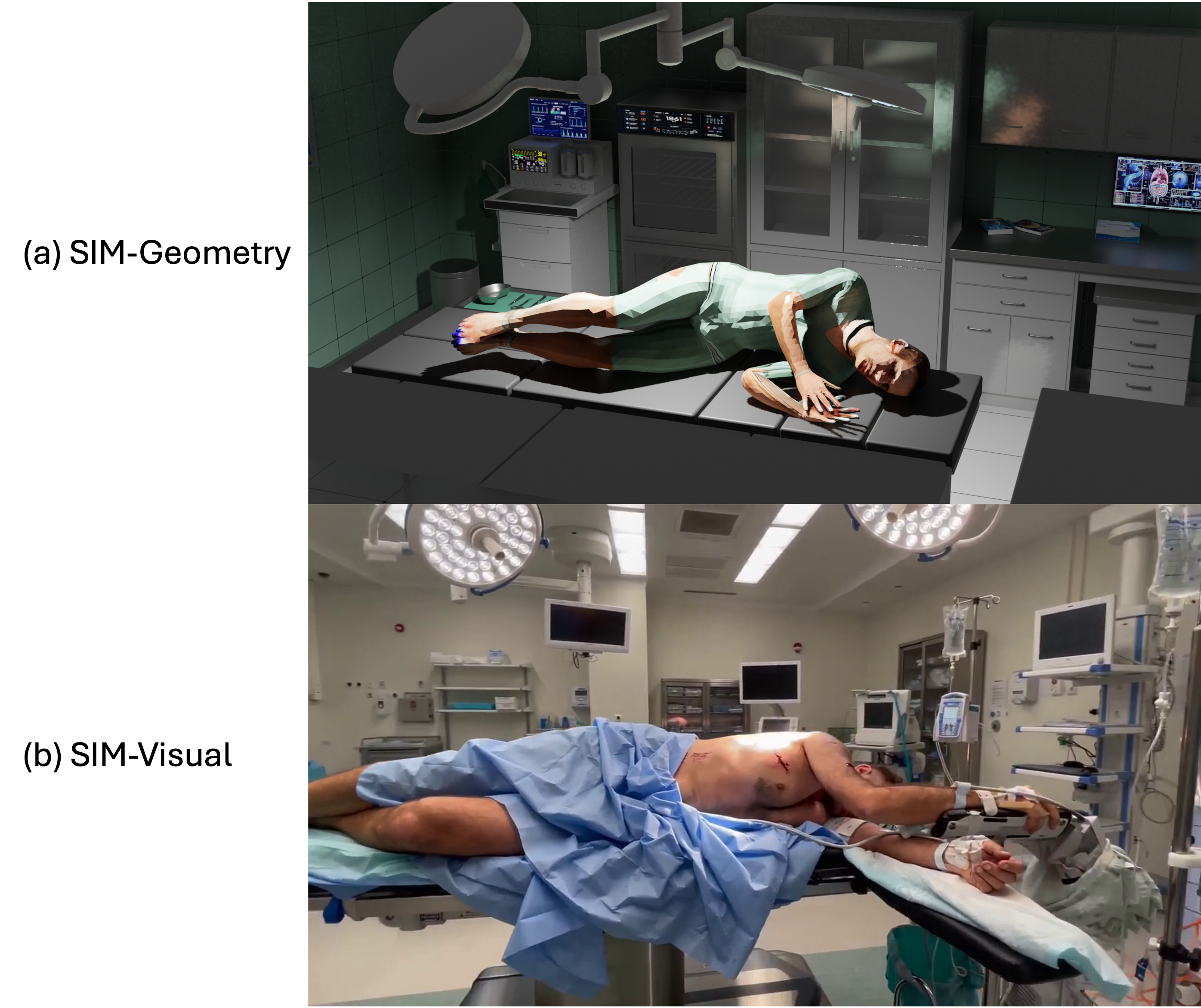}
  \caption{Representative frames from the simulation datasets. Top: Sim-Geometry sequences rendered from SLP-3Dfits with ground truth SMPL meshes, generated patient texture and orbiting camera silhouettes. Bottom: Sim-Visual frames at three drape coverage levels (none, partial, heavy), depicting realistic surgical scenes with photorealistic skin textures and operating room environments. These two are complementary: Sim-Geometry shows clean silhouettes with ground truth SMPL meshes whereas Sim-Visual demonstrates increasing visual complexity from drapes.}
  \label{fig:sim_dataset}
\end{figure}

\textbf{Temporal chaining.}
Frames were processed in forward and backward passes from the initial keyframe, each initialized from the nearest fitted neighbor. A moving-average filter smoothed the resulting trajectories. Temporal regularization was disabled when the gap to the nearest fitted frame exceeded a gap threshold $\tau_g$, accommodating large camera movements between distant observations.

Rigid Fallback operated in two modes: \textit{full} mode replaced all frames with rigid fitting, and \textit{fallback} mode selectively replaced only frames where the per-frame Mesh-Mask IoU fell below a threshold $\tau_\text{IoU}$. We used fallback mode ($\tau_\text{IoU}{=}0.6$) as the default: this preserved high-quality per-frame HMR predictions on well-observed frames while correcting failures on frames with limited body visibility.

\subsection{Application to Surgical AR}
\label{sec:application}

The recovered SMPL meshes can serve as geometric proxies for registering preoperative imaging (CT/MRI) onto the surgical scene. Given a preoperative 3D model (e.g., from CT), the SMPL surface provides a target for non-rigid registration, aligning the internal anatomy to the observed body surface in each frame. Prior work has shown that the SMPL body surface can predict internal skeletal shape~\cite{osso}, suggesting that recovered meshes could also serve as input for anatomical inference. However, SMPL captures statistical body shape variation and does not model patient-specific anatomical features. The recovered surface therefore provides an approximate registration target; sub-centimeter clinical accuracy would require additional refinement using intraoperative depth data or deformable registration constrained by preoperative imaging. Because Patient4D produces temporally consistent meshes, the registration only needs to be computed once and can be propagated across the video via the per-frame rigid transformations $(\mathbf{R}_t, \mathbf{c}_t)$ from the Rigid Fallback stage. The primary remaining challenge is depth ambiguity, discussed in \cref{sec:discussion}.


\section{Experiments}
\label{sec:experiments}

Our evaluation addresses two questions: (1) does Patient4D outperform existing HMR methods on surgical video, and (2) does it retain competitive accuracy on standard HMR benchmarks? While real operating room videos with 3D ground truth are unavailable due to privacy and sterility constraints, synthetic surgical datasets have been widely used to benchmark vision algorithms in clinical environments~\cite{4dor,mmor}.

\subsection{Setup}
\label{sec:exp_setup}

\textbf{Simulation datasets.}
We constructed two complementary simulation datasets that isolate different evaluation dimensions (\cref{tab:sim_dataset}, \cref{fig:sim_dataset}). Sim-Geometry ensures geometric correctness while Sim-Visual stresses visual complexity and occlusion patterns. No simulation videos use real patient data.

\textit{Sim-Geometry.} To evaluate with 3D ground truth, we generated synthetic sequences from the SLP-3Dfits dataset~\cite{slp3dfits}, which provides SMPL fits for 4,545 lying-down poses. For each pose, a virtual camera orbited 360$^\circ$ around the body and rendered silhouette masks with pyrender. Ground truth includes per-frame SMPL vertices, camera extrinsics, and silhouette masks, and the body is textured using~\cite{tu2025smpl}. Sim-Geometry provides an upper bound on pipeline accuracy in the absence of visual complexity: no drapes, textured bodies, clean backgrounds.

\textit{Sim-Visual.} To test the pipeline under realistic visual conditions without clinical 3D ground truth, we generated surgical scene videos using Seedance2.0~\cite{seedance2_0}, a text-to-video model. The use of synthetic data for evaluating human mesh recovery is well established: BEDLAM~\cite{bedlam} demonstrated that models trained entirely on rendered synthetic humans achieve state-of-the-art accuracy on real-image benchmarks, and prior OR scene understanding datasets such as 4D-OR~\cite{4dor} and MM-OR~\cite{mmor} rely on simulated surgical procedures with non-medical actors rather than real patients. We designed 135 text prompts that systematically varied patient pose (supine, lateral decubitus, prone), drape coverage level (none, partial, heavy), and camera trajectory ($120^\circ$--$180^\circ$ orbits, lateral pans, crane shots, first-person surgeon perspectives). Full prompt specifications are provided in supplementary material.

To assess the clinical fidelity of Sim-Visual, two surgeons reviewed a subset of 10 sequences using a structured five-point Likert checklist covering operating room realism, anatomical proportions, body position appropriateness, and drape pattern plausibility. The mean realism score across all questions was 4.2 out of 5 (``Mostly realistic''), and all 10 videos were accepted as valid proxies for evaluating body estimation algorithms. Minor differences from real surgical footage included the absence of monitoring lines (IV, blood pressure cuff) and the use of non-surgical beds in some sequences. The full checklist and per-video results are provided in supplementary material.

\begin{figure*}[!htbp]
  \centering
  \includegraphics[width=\textwidth, height=0.92\textheight, keepaspectratio]{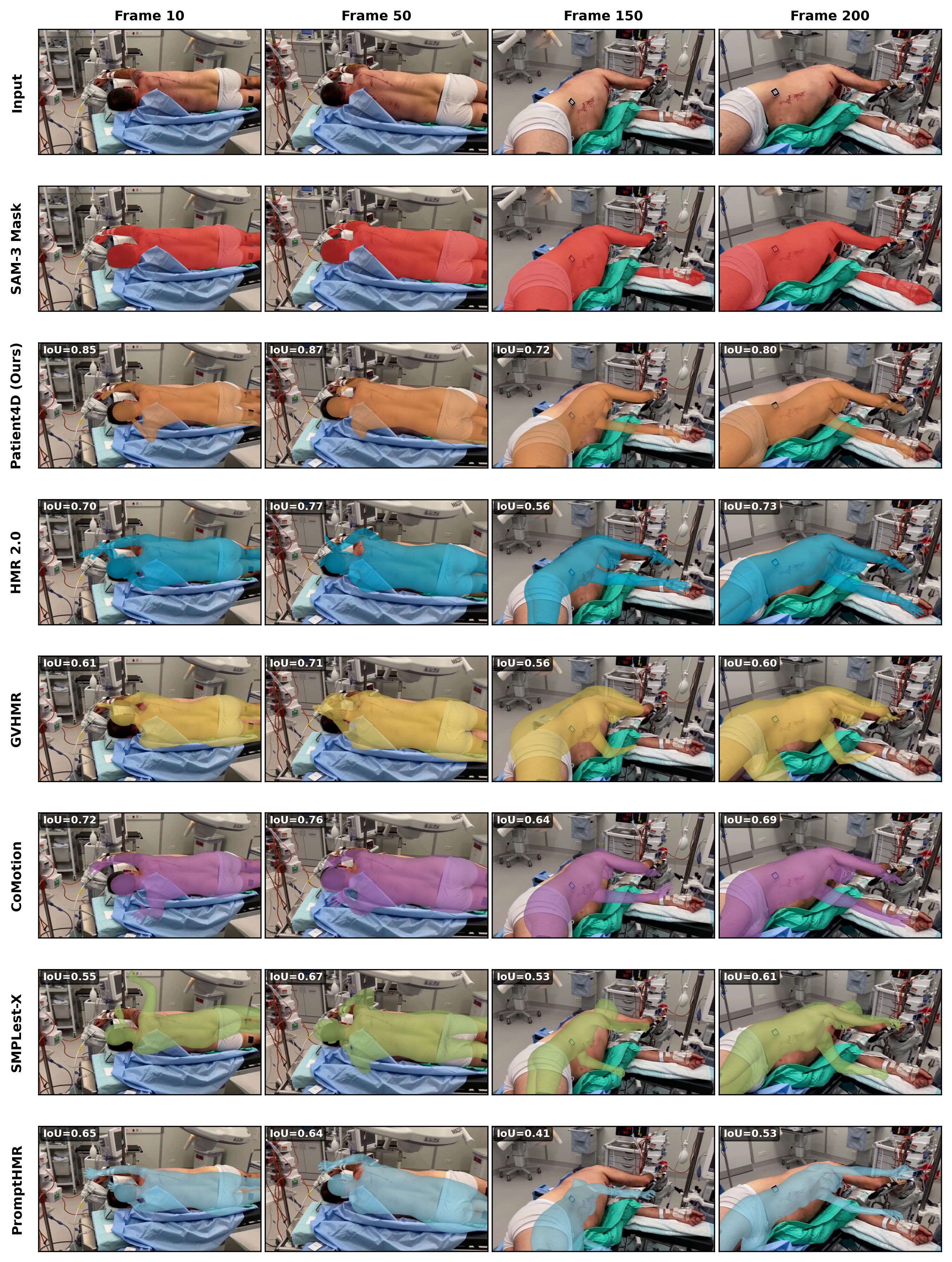}
  \caption{Qualitative comparison of mesh recovery on a synthetic surgical sequence. The patient is positioned in a lateral decubitus pose under no draping. As the camera orbits from the initial viewpoint (Frames 10, 50) to extreme oblique angles (Frames 150, 200), the 2D visual cues become ambiguous. Standard per-frame and video-based baselines fail to recognize the stationarity prior of the patient; they attempt to fit the changing 2D contours by distorting the 3D anatomy (e.g., unnaturally twisting the spine or flattening the torso), resulting in plummeting Mesh-Mask IoU. In contrast, Patient4D (orange) enforces temporal consistency, preserving the correct 3D geometry and maintaining strong spatial alignment across the entire camera trajectory.}
  \label{fig:qualitative}
\end{figure*}

\begin{table}[!tb]
\centering
\caption{Simulation dataset configurations.}
\label{tab:sim_dataset}
\begin{tabular}{lll}
\toprule
\textbf{Parameter} & \textbf{Sim-Geometry} & \textbf{Sim-Visual} \\
\midrule
Source & SLP-3Dfits~\cite{slp3dfits} & Seedance2.0~\cite{seedance2_0} \\
Sequences & 4545 & 135 \\
Frames/seq. & 375 (25\,fps) & ${\sim}$192 (24\,fps) \\
Resolution & 760 $\times$ 428 & 1280 $\times$ 720 \\
3D ground truth & Yes & No \\
Drape occlusion & None & 3 levels \\
\bottomrule
\end{tabular}
\end{table}

\textbf{Standard HMR Video Benchmarks.}
We also evaluated on 3DPW~\cite{3dpw} (standard test split), EMDB~\cite{emdb}, and RICH~\cite{rich} to demonstrate that Patient4D retains state-of-the-art accuracy on standard HMR alongside its surgical capabilities. 3DPW features upright, moving subjects with IMU-augmented video ground truth. EMDB~\cite{emdb} provides per-frame SMPL ground truth derived from electromagnetic body sensors across 17 outdoor sequences, covering a broader range of activities including athletic movements and offering more accurate reference poses than marker-based capture. RICH provides per-frame SMPL-X ground truth in non-standard poses including pushups and bending. 

\textbf{Metrics.}
On Sim-Geometry, where 3D ground truth SMPL meshes are available, we report Procrustes-aligned per-vertex error (PA-PVE, in mm) to measure 3D reconstruction accuracy. Because the pipeline's multi-resolution mesh (18,439 vertices) differs in topology from the SMPL ground truth (6,890 vertices), PA-PVE was computed via Procrustes alignment followed by nearest-neighbor vertex matching, which may introduce minor noise in the vertex-level metric.

On Sim-Visual videos, where no 3D ground truth exists, we rely on 2D metrics: \textit{Mesh-Mask IoU} between the projected mesh silhouette and the SAM3 mask, and \textit{temporal stability} $\bar{\Delta}_{\text{mesh}} = \frac{1}{T-1}\sum_{t=1}^{T-1}\frac{1}{N_v}\|\mathbf{V}_t - \mathbf{V}_{t-1}\|_F$ (Frobenius-normalized per-vertex displacement between consecutive frames, in meters). We also report the percentage of frames above 0.6 IoU and below 0.3 IoU. We acknowledge that 2D IoU cannot fully characterize 3D mesh accuracy. However, Sim-Geometry provides complementary 3D evaluation (PA-PVE), and Patient4D achieves the best result on both metrics (0.75 IoU and 78.3mm PA-PVE), indicating that improvements in silhouette alignment are consistent with improvements in 3D reconstruction. To comprehensively evaluate 3D structural correctness on Sim-Visual, we introduce two metrics tailored for the OR environment: 
(1) \textit{Pose Consistency ($\Delta_{\text{pose}}$)}: the mean L2 distance of SMPL pose parameters between consecutive frames, defined as $\frac{1}{T-1} \sum_{t=1}^{T-1} \|\boldsymbol{\theta}_t - \boldsymbol{\theta}_{t-1}\|_2$ (in radians). This isolates internal articulation stability from global camera motion. 
(2) \textit{Cross-View IoU ($\text{CV-IoU}_{20}$)}: the IoU obtained by projecting the 3D mesh from frame $t$ into the camera viewpoint of frame $t+20$, defined as $\frac{1}{T-20} \sum_{t=1}^{T-20} \text{IoU}(\text{Project}(\mathbf{V}_t, \mathbf{K}, \mathbf{R}_{t+20}, \mathbf{c}_{t+20}), S_{t+20})$. Since the patient is static, a geometrically correct 3D mesh should accurately align with the future mask despite the camera orbit. This metric severely penalizes methods that unnaturally flatten or contort the 3D pose to match a single 2D perspective.

On 3DPW and EMDB, we report PA-MPJPE, MPJPE, and PVE (all in mm) following standard protocol. On RICH, we report PA-PVE via nearest-neighbor vertex matching. Statistical significance between Patient4D and the best-performing baseline on the simulation datasets was assessed using a paired Wilcoxon signed-rank test, with $p < 0.05$ considered significant.

\textbf{Baselines.}
We compare against five state-of-the-art methods using their official implementations: HMR 2.0~\cite{hmr2} and SMPLest-X~\cite{smplesttx} (per-frame); GVHMR~\cite{gvhmr} and CoMotion~\cite{comotion} (video-based); and PromptHMR~\cite{prompthmr} (mask-promptable, per-frame). PromptHMR accepts segmentation masks as spatial prompts, so we provide it with the same SAM3 masks used by Patient4D, ensuring a fair comparison of the downstream mesh recovery.

\textbf{Implementation.}
All experiments use a single NVIDIA RTX 3090 GPU (24~GB VRAM). Patient4D uses Rigid Fallback in fallback mode: frames with Mesh-Mask IoU below 0.6 are replaced by rigidly fitting the best reference from the keyframe pool ($\tau_q{=}0.6$, $\tau_d{=}50$ frames) via Nelder-Mead simplex optimization~\cite{nelder1965simplex} at half resolution (150 iterations, $\lambda_\text{temp}{=}0.1$). Shape $\beta$ and scale are locked to keyframe-pool values. Post-processing hyperparameters include: outlier thresholds $\tau^\theta{=}0.6$, $\tau^r{=}0.6$, $\tau^{\mathbf{c}}{=}0.2$; median window $w{=}7$; Pose Locking warmup $W{=}48$, anchor pull $\alpha_a{=}0.005$.

\subsection{Simulation Results and Analysis}

\textbf{Sim-Geometry.}
On Sim-Geometry (\cref{tab:sim_baselines}), Patient4D achieved 0.75 Mesh-Mask IoU and 78.3~mm PA-PVE, outperforming all baselines (next best: HMR~2.0 at 0.68 IoU, 84.2~mm PA-PVE). Patient4D produced zero failure frames ($<$0.3 IoU) compared to 4.5--32.5\% for baselines, and the lowest temporal displacement (0.008~m). Since Sim-Geometry presents the full body without occlusion, most frames exceeded the Rigid Fallback threshold ($\tau_\text{IoU}{=}0.6$), and the output largely reflected raw per-frame HMR quality. The 78.3~mm PA-PVE captures both genuine pose estimation error and the topology mismatch between the 18,439-vertex pipeline mesh and the 6,890-vertex SMPL ground truth.

\textbf{Sim-Visual.}
On Sim-Visual (\cref{tab:sim_baselines}), Patient4D achieved 0.75 Mesh-Mask IoU, significantly outperforming all baselines (next best: HMR 2.0 at 0.52, $p < 0.001$). While \cref{tab:sim_baselines} establishes Patient4D's overall superiority against baselines, \cref{tab:sim_visual_results} provides a dedicated stress test of Patient4D by breaking down its performance across varying drape levels. IoU decreased with increasing drape coverage: the no-drape condition achieved the highest IoU (0.84), followed by partial draping (0.73) and heavy draping (0.69 IoU, 3.3\% failure frames), where the visible area provides insufficient geometric signal. \cref{fig:qualitative} visualizes this advantage, showing how Patient4D maintains accurate mesh alignment under severe camera displacement while baselines progressively drift. Across all drape levels, the performance gap over baselines was consistent and statistically significant: HMR~2.0 achieved 0.52 overall IoU with a 35.1\% failure rate versus Patient4D's 0.75 with only a 1.3\% failure rate ($p < 0.001$).

The cross-view and pose consistency metrics confirm that Patient4D recovers true 3D volumetric geometry. As shown in \cref{tab:sim_baselines}, baselines degrade on Cross-View IoU ($\text{CV-IoU}_{20}$), dropping to 0.28 for HMR 2.0. This indicates that their per-frame 3D distortions fail to generalize when projected from novel viewpoints. In contrast, Patient4D maintains a strong $\text{CV-IoU}_{20}$ of 0.68, proving the recovered 3D anatomy is geometrically correct beneath the drapes. Furthermore, Patient4D achieves a near-zero Pose Consistency error ($\Delta_{\text{pose}} = 0.01$ rad), confirming that the internal articulation remains rigidly locked across the sequence, overcoming the erratic joint jitter seen in standard detection-based methods.

\begin{table}[!tb]
\centering
\caption{Performance breakdown of Patient4D across varying levels of drape occlusion on the Sim-Visual dataset. While overall baseline comparisons are provided in \cref{tab:sim_baselines}, this analysis isolates how Patient4D degrades under increasing visual ambiguity. IoU: Mesh-Mask IoU ($\uparrow$). $\bar{\Delta}$: mean vertex displacement ($\downarrow$).}
\label{tab:sim_visual_results}
\begin{tabular}{@{}lccccc@{}}
\toprule
Drape Level & \textbf{IoU}$\uparrow$ & $>$0.6$\uparrow$ & $<$0.3$\downarrow$ & $\bar{\Delta}$$\downarrow$ & \textbf{Seqs} \\
\midrule
None    & 0.84 & 99.5\% & 0.0\% & 0.010 & 30 \\
Partial & 0.73 & 94.5\% & 0.5\% & 0.051 & 60 \\
Heavy   & 0.69 & 80.1\% & 3.3\% & 0.013 & 45 \\
\midrule
All     & 0.75 & 90.8\% & 1.3\% & 0.029 & 135 \\
\bottomrule
\end{tabular}
\end{table}

\begin{table*}[!t]
\centering
\caption{Comparison on simulation datasets. Sim-Geometry isolates pose complexity with 3D ground truth. Sim-Visual isolates occlusion with realistic rendering. Best results are in \textbf{bold}. The asterisk ($^*$) indicates a statistically significant improvement ($p < 0.001$).}
\label{tab:sim_baselines}
\setlength{\tabcolsep}{5pt}
\begin{tabular}{l|ccccc|ccccc}
\toprule
& \multicolumn{5}{c|}{\textbf{Sim-Geometry}} & \multicolumn{5}{c}{\textbf{Sim-Visual}} \\
\textbf{Method} & PA-PVE$\downarrow$ & IoU$\uparrow$ & $>$0.6$\uparrow$ & $<$0.3$\downarrow$ & $\bar{\Delta}$$\downarrow$ & IoU$\uparrow$ & CV-IoU$_{20}\uparrow$ & $<$0.3$\downarrow$ & $\Delta_{\text{pose}}$$\downarrow$ & $\bar{\Delta}$$\downarrow$ \\
\midrule
HMR 2.0~\cite{hmr2} & 84.2 & 0.68 & 78.2 & 4.5 & 0.185 & 0.52 & 0.28 & 35.1 & 0.14 & 0.650 \\
GVHMR~\cite{gvhmr} & 95.6 & 0.58 & 62.4 & 8.2 & 0.014 & 0.45 & 0.31 & 30.5 & 0.05 & \textbf{0.025} \\
CoMotion~\cite{comotion} & 87.9 & 0.65 & 71.5 & 10.1 & 0.042 & 0.38 & 0.22 & 52.4 & 0.08 & 0.065 \\
SMPLest-X~\cite{smplesttx} & 112.4 & 0.48 & 35.2 & 32.5 & 1.850 & 0.22 & 0.11 & 70.1 & 0.45 & 2.450 \\
PromptHMR~\cite{prompthmr} & 89.5 & 0.62 & 68.3 & 9.4 & 0.135 & 0.35 & 0.19 & 45.2 & 0.11 & 0.140 \\
\midrule
Patient4D (Ours) & \textbf{78.3} & \textbf{0.75} & \textbf{94.5} & \textbf{0.0} & \textbf{0.008} & \textbf{0.75}$^*$ & \textbf{0.68}$^*$ & \textbf{1.3}$^*$ & \textbf{0.01}$^*$ & 0.029 \\
\bottomrule
\end{tabular}
\end{table*}

\begin{table}[!tb]
\centering
\caption{Results on standard HMR video benchmarks. 3DPW and EMDB: camera-space metrics (PA-MPJPE / MPJPE / PVE, mm).
RICH: PA-PVE (mm) via nearest-neighbor vertex matching. The best result in each column is in \textbf{bold}.}
\label{tab:benchmarks}
\begin{tabular}{lccc}
\toprule
\multicolumn{4}{c}{\textbf{3DPW}} \\
\midrule
\textbf{Method} & \textbf{PA-MPJPE}$\downarrow$ & \textbf{MPJPE}$\downarrow$ & \textbf{PVE}$\downarrow$ \\
\midrule
HMR 2.0~\cite{hmr2} & 44.4 & 69.8 & 82.2 \\
GVHMR~\cite{gvhmr} & 37.0 & 56.6 & 68.7 \\
CoMotion~\cite{comotion} & 37.3 & 60.0 & 71.5 \\
SMPLest-X~\cite{smplesttx} & 45.2 & 74.8 & 88.4 \\
PromptHMR~\cite{prompthmr} & 36.6 & 58.7 & 69.4 \\
Patient4D (Ours) & \textbf{35.8} & \textbf{54.8} & \textbf{65.2} \\
\midrule
\multicolumn{4}{c}{\textbf{EMDB}} \\
\midrule
\textbf{Method} & \textbf{PA-MPJPE}$\downarrow$ & \textbf{MPJPE}$\downarrow$ & \textbf{PVE}$\downarrow$ \\
\midrule
HMR 2.0~\cite{hmr2} & 60.6 & 98.0 & 120.3 \\
GVHMR~\cite{gvhmr} & 42.7 & 72.6 & 84.9 \\
CoMotion~\cite{comotion} & 43.1 & 73.5 & 86.2 \\
SMPLest-X~\cite{smplesttx} & 58.4 & 92.1 & 115.5 \\
PromptHMR~\cite{prompthmr} & \textbf{41.0} & 71.7 & \textbf{84.5} \\
Patient4D (Ours) & 41.3 & \textbf{71.5} & 84.7 \\
\midrule
\multicolumn{4}{c}{\textbf{RICH}} \\
\midrule
\textbf{Method} & \multicolumn{3}{c}{\textbf{PA-PVE}$\downarrow$} \\
\midrule
HMR 2.0~\cite{hmr2} & \multicolumn{3}{c}{116.2} \\
GVHMR~\cite{gvhmr} & \multicolumn{3}{c}{121.4} \\
CoMotion~\cite{comotion} & \multicolumn{3}{c}{128.7} \\
SMPLest-X~\cite{smplesttx} & \multicolumn{3}{c}{142.3} \\
PromptHMR~\cite{prompthmr} & \multicolumn{3}{c}{108.5} \\
Patient4D (Ours) & \multicolumn{3}{c}{\textbf{99.8}} \\
\bottomrule
\end{tabular}
\end{table}

\begin{figure*}[!tb]
  \centering
  \includegraphics[width=\linewidth]{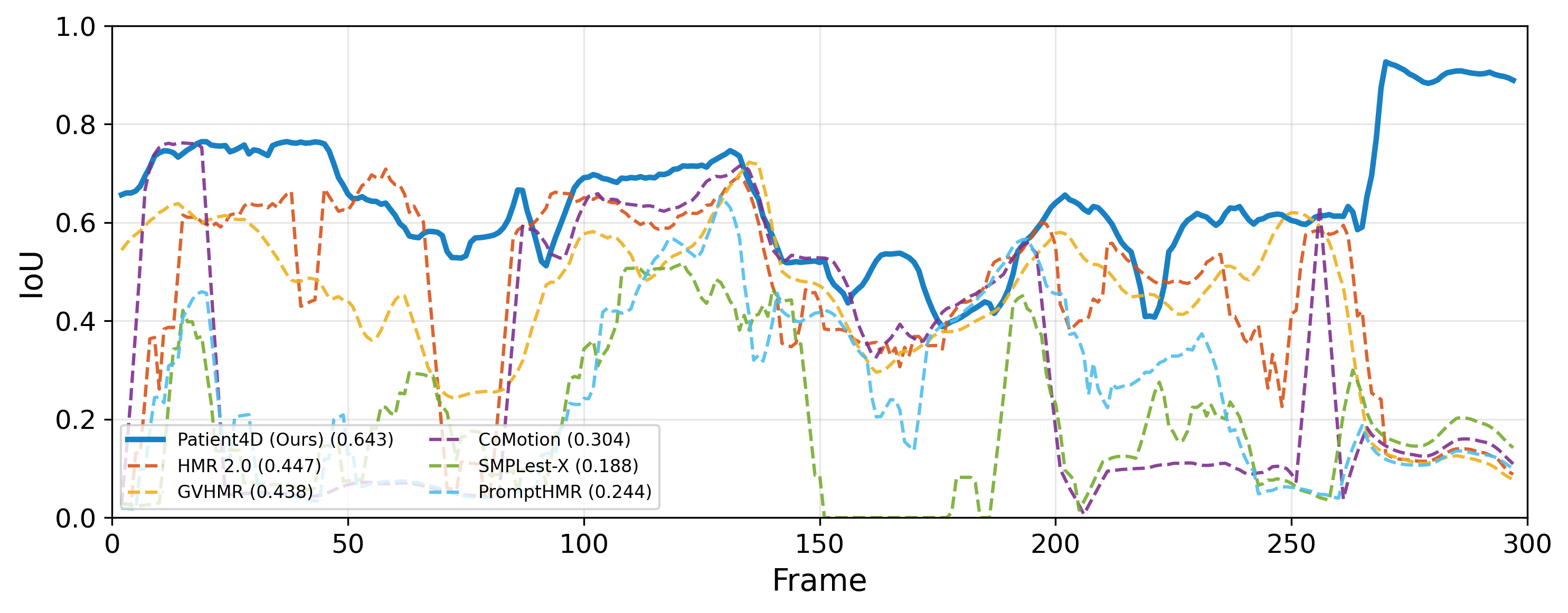}
  \caption{Per-frame Mesh-Mask IoU ($\uparrow$) over time for one simulation video. At approximately frame 260, the sequence introduces a severe camera viewpoint transition coupled with heavy drape occlusion. While all baseline methods suffer failure frames (IoU dropping below 0.2), Patient4D successfully triggers Rigid Fallback, retrieving a high-quality historical reference to immediately recover and maintain high spatial alignment (IoU $\approx$ 0.9).}
  \label{fig:temporal}
\end{figure*}

\begin{figure*}[!tb]
  \centering
  \includegraphics[width=\linewidth]{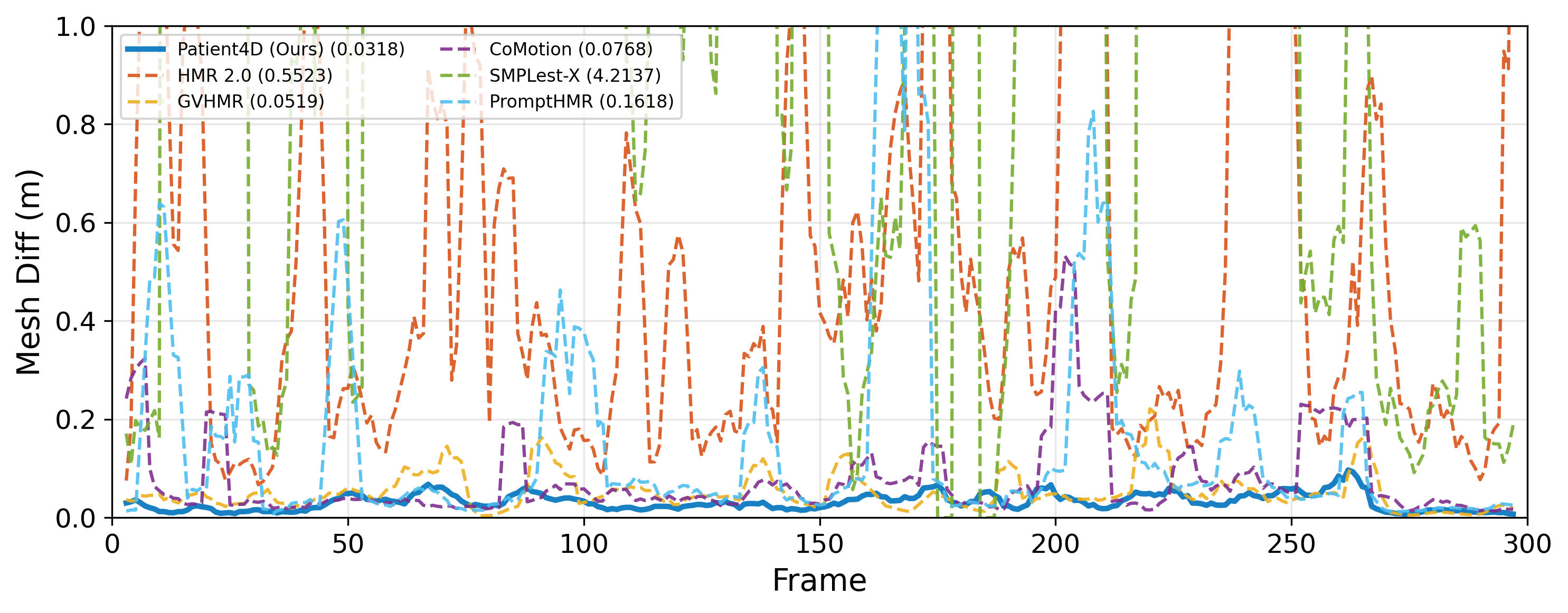}
  \caption{Per-frame mesh vertex displacement ($\downarrow$) between consecutive frames. Detection-based baselines (HMR 2.0, SMPLest-X, PromptHMR) exhibit large inter-frame jumps when detections shift or fail, with SMPLest-X exceeding 1\,m regularly (capped for visibility). Patient4D and GVHMR maintain low displacement ($<$0.1\,m), reflecting temporally consistent reconstructions.}
  \label{fig:temporal_mesh}
\end{figure*}

\cref{fig:temporal} shows temporal IoU progression across all methods, demonstrating how Patient4D maintains consistently high alignment while baselines fluctuate. Furthermore, \cref{fig:temporal_mesh} highlights the severe inter-frame instability ($\bar{\Delta}$) of detection-based baselines compared to our temporally consistent reconstructions.

\subsection{Evaluation on Standard HMR Video Benchmarks}

\begin{figure}[!tb]
  \centering
  \includegraphics[width=\linewidth]{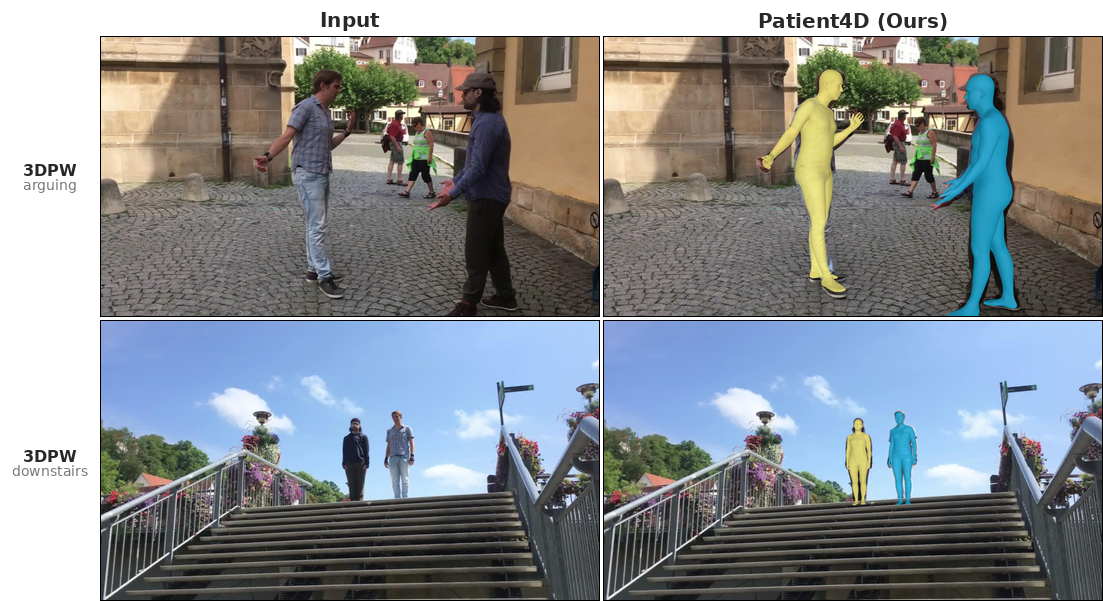}
  \caption{Qualitative results on 3DPW video sequences. Patient4D recovers accurate meshes for multiple people in outdoor scenes without any domain-specific tuning. Each detected person is shown in a distinct color. Top: two people arguing on a cobblestone street. Bottom: two people descending stairs.}                        
  \label{fig:benchmark_qual}
\end{figure}

On 3DPW (\cref{tab:benchmarks}, \cref{fig:benchmark_qual}), Patient4D achieved 35.8~mm PA-MPJPE, 54.8~mm MPJPE, and 65.2~mm PVE, the best results across all methods including video-based approaches (GVHMR, CoMotion). Patient4D achieves these results without modifying the HMR backbone. The improvement over baselines stems from the mask-conditioned input provided by SAM3 and the stable focal length estimation from MoGe. Because the surgical temporal modules are disabled on these benchmarks, the pipeline preserves the zero-shot accuracy of the underlying foundation models.
 
On EMDB, Patient4D achieved 41.3~mm PA-MPJPE and 71.5~mm MPJPE, comparable to the best method (PromptHMR: 41.0 / 71.7~mm). On RICH, Patient4D achieved 99.8~mm PA-PVE, the best among all methods.

\subsection{Ablation Study}
 
\cref{tab:ablation} compares an incremental post-processing chain (rows A--E) with Patient4D's final configuration (row F), evaluated on simulation data.

\begin{table}[!tb]
\centering
\caption{Component ablation (macro-averaged across Sim-Geometry and Sim-Visual datasets). Rows A--E: sequential additive chain. Row F (Patient4D): skips smoothing and applies Pose Locking and selective Rigid Fallback directly to outlier-rejected raw predictions.}
\label{tab:ablation}
\setlength{\tabcolsep}{4pt}
\begin{tabular}{@{}lcccc@{}}
\toprule
\textbf{Config.} & \textbf{IoU}$\uparrow$ & \textbf{$>$0.6}$\uparrow$ & \textbf{$<$0.3}$\downarrow$ & $\bar{\Delta}$$\downarrow$ \\
\midrule
(A) SAM3 + SAM3DBody (raw) & 0.70 & 84.0 & 2.1 & 0.024 \\
(B) (A) + Outlier rejection & 0.71 & 84.2 & 2.3 & 0.022 \\
(C) (B) + Smoothing & 0.58 & 60.5 & 8.5 & 0.022 \\
(D) (C) + Pose Locking & 0.71 & 84.5 & 1.8 & 0.022 \\
(E) (D) + Rigid Fallback (full) & 0.72 & 85.0 & 2.0 & 0.022 \\
\midrule
(F) \textbf{Patient4D (Ours)} & \textbf{0.75} & \textbf{92.7} & \textbf{0.7} & 0.019 \\
\bottomrule
\end{tabular}
\end{table}

Outlier rejection (B) marginally increased IoU from 0.70 to 0.71. Smoothing (C) produced the largest degradation ($-$0.13 IoU) because temporal averaging blends correct predictions from well-observed frames with incorrect predictions from occluded frames, pulling both toward an inaccurate mean. Pose Locking (D) recovered from this degradation and slightly exceeded raw accuracy, confirming that anchoring body pose to a stable median is beneficial even when earlier smoothing stages are harmful. The full additive chain with rigid fitting (E) recovered most of the lost accuracy but started from smoothed inputs, which provided a worse starting point for rigid optimization.

Patient4D (F) achieved the best overall results: 0.75 IoU, 92.7\% of frames above the 0.6 IoU threshold, and only 0.7\% failure frames. By skipping smoothing and applying Pose Locking and selective Rigid Fallback directly to outlier-rejected raw predictions, it preserved high-quality predictions on well-observed frames (84.0\% exceeded 0.6 IoU in row A) while correcting failures on difficult frames. The mesh displacement ($\bar{\Delta} = 0.019$ m) was also the lowest among all configurations.

\begin{table}[!tb]
\centering
\caption{Sensitivity of Rigid Fallback to temporal regularization $\lambda_\text{temp}$ and optimization iterations, averaged across all surgical sequences.}
\label{tab:sensitivity}
\begin{tabular}{lccc}
\toprule
\textbf{Parameter} & \textbf{Value} & \textbf{IoU}$\uparrow$ & $\bar{\Delta}$$\downarrow$ \\
\midrule
\multirow{4}{*}{$\lambda_\text{temp}$}
& 0 (none) & 0.75 & 0.021 \\
& 0.05 & 0.75 & 0.020 \\
& \textbf{0.1 (ours)} & \textbf{0.75} & \textbf{0.019} \\
& 0.5 & 0.73 & 0.018 \\
\midrule
\multirow{3}{*}{Max iterations}
& 50 & 0.74 & 0.020 \\
& \textbf{150 (ours)} & \textbf{0.75} & \textbf{0.019} \\
& 300 & 0.75 & 0.019 \\
\bottomrule
\end{tabular}
\end{table}

\subsection{Computational Cost}
 
\cref{tab:efficiency} reports per-stage runtime for Patient4D on a 300-frame sequence (1280$\times$720, 24~fps). All components are either pre-trained foundation models or optimization-based, incurring zero training cost. The total 638-second runtime is offline-only, but it compares favorably to the preoperative setup required by marker-based registration systems, which involves attaching fiducials, aligning, and verifying, a process that can take 8--10 minutes per patient~\cite{ribmr}.

\begin{table}[!tb]
\centering
\caption{Per-stage cost for 300 frames on a single NVIDIA RTX 3090 GPU.}
\label{tab:efficiency}
\begin{tabular}{lccc}
\toprule
\textbf{Stage} & \textbf{Time (s)} & \textbf{VRAM (GB)} & \textbf{Device} \\
\midrule
SAM3 segmentation & 40 & 12.4 & GPU \\
MoGe FOV & 2 & 3.8 & GPU \\
SAM3DBody HMR & 550 & 18.2 & GPU \\
Post-processing & $<$1 & -- & CPU \\
Rigid fallback & 45 & 1.2 & GPU \\
\midrule
\textbf{Total} & \textbf{638} & \textbf{18.2} & -- \\
\bottomrule
\end{tabular}
\end{table}

\section{Discussion}
\label{sec:discussion}
 
Patient4D outperformed all baselines on surgical simulation, achieving 0.75 mean IoU against 0.52 for HMR~2.0 and reducing failure frames to 1.3\%. On standard HMR video benchmarks, the pipeline retained competitive accuracy with its surgical modules disabled. The ablation confirmed that Pose Locking and selective Rigid Fallback, applied to raw per-frame predictions without smoothing, outperformed the full additive post-processing chain.
 
\textbf{The stationarity prior as a hard constraint.}
The ablation (\cref{tab:ablation}) reveals that temporal smoothing is counterproductive in this domain: averaging correct predictions with incorrect predictions on occluded frames degrades both. Pose Locking recovers from this degradation by anchoring body pose to a stable median, confirming that the static patient assumption is valuable when encoded as a hard geometric constraint rather than a soft temporal average. Rigid Fallback complements this by discarding unreliable predictions entirely and substituting a geometrically grounded alternative from the keyframe pool. This observation suggests that when a strong domain prior is available, encoding it as a hard constraint can outperform learned temporal priors trained on a different distribution.
 
\textbf{Baseline failure analysis.}
HMR~2.0 and SMPLest-X, as per-frame methods trained on BEDLAM and AMASS, produced predictions that lacked temporal consistency. SMPLest-X exhibited the highest instability, likely due to its per-bounding-box focal length estimation, which produces inconsistent scale across frames. GVHMR achieved good temporal stability ($\bar{\Delta}{=}0.025$) through gravity-aligned coordinates, but this coordinate system assumes vertical subjects and penalizes supine or lateral poses. CoMotion, designed for multi-person tracking via Hungarian assignment, performed poorly on single static subjects. PromptHMR, despite receiving the same SAM3 masks as Patient4D, still struggled with non-standard poses. This confirms that spatial mask guidance alone is not sufficient when the underlying HMR backbone lacks a domain-appropriate prior. The $\text{CV-IoU}_{20}$ metric reveals whether methods achieve true 3D reconstruction or merely overfit 2D silhouettes via implausible $Z$-axis compression. Patient4D’s strong $\text{CV-IoU}_{20}$ and minimal $\Delta_{\text{pose}}$ prove our static prior successfully prevents such view-dependent distortions, ensuring globally coherent 3D meshes.
 
\textbf{Generalizability.}
The principle of encoding a stationarity prior as a hard constraint may extend to other settings where subjects remain approximately static: sleep studies, rehabilitation monitoring, patient positioning verification, or long-duration imaging sessions. In each case, standard video-based HMR methods that assume subject motion will underperform, and a rigid-body assumption similar to Rigid Fallback could be applied. 
 
\textbf{Limitations.}
The pipeline has several limitations. First, Rigid Fallback depends on the availability of high-quality keyframes. If most frames are heavily occluded and the keyframe pool does not contain a good prediction, the reference mesh itself may be inaccurate. Second, the static patient assumption breaks down for awake patients or intraoperative repositioning events. Monocular RGB also cannot recover absolute depth, which limits registration accuracy along the camera axis. Fusing metric depth from the headset's time-of-flight sensor or calibrating against a known physical reference in the scene would resolve this. The Rigid Fallback depth regularization (\cref{eq:rigid_objective}) already maintains relative depth consistency across frames, providing a foundation for such integration. Collecting RGBD ground truth in clinical settings is an important next step for more comprehensive evaluation.
 
As future work, we plan to address these limitations directly. Fusing metric depth from headset depth sensors would resolve the depth ambiguity. Integrating the recovered meshes into an end-to-end surgical AR registration pipeline with downstream navigation accuracy evaluation would validate the practical clinical value. Future work includes clinical evaluation on real surgical footage under appropriate institutional approvals.

\section{Conclusion}
\label{sec:conclusion}
 
We have presented Patient4D, a stationarity-constrained pipeline for temporally consistent 4D body mesh recovery from monocular surgical video. Patient4D achieves state-of-the-art results on surgical simulation datasets while retaining competitive accuracy on 3DPW, EMDB, and RICH, without additional training or domain-specific data. Pose Locking anchors body pose predictions to a robust median estimate, reducing temporal jitter under shifting viewpoints. Rigid Fallback corrected unreliable predictions on difficult frames by rigidly transforming a high-quality reference mesh from the keyframe pool, reducing failure frames to 1.3\% compared to 30.5\% for the best baseline. The ablation confirmed that these two geometric mechanisms that operationalize the stationarity constraint, applied to raw per-frame predictions without smoothing, outperformed a full incremental post-processing chain. The recovered SMPL meshes can serve as geometric proxies for registering preoperative imaging onto the surgical scene. As future work, resolving depth ambiguity through metric depth sensors, replacing Nelder-Mead with differentiable rendering for faster rigid fitting, and clinical validation with RGBD ground truth across diverse procedures would further strengthen the pipeline's practical utility.

\section{Acknowledgments}
The Sim-Visual dataset was generated using Seedance2.0~\cite{seedance2_0}, a text-to-video generative AI model. Text prompts were designed by the authors; generated videos were manually curated for quality. Gemini 3.0 Pro was used in revising all sections and correcting grammar.
\bibliographystyle{abbrv-doi}
\bibliography{template}

\end{document}